# Syntax-Semantics Interaction Parsing Strategies. Inside SYNTAGMA

Daniel Christen
(www.lector.ch)

**Abstract**

This paper discusses SYNTAGMA, a rule based NLP system addressing the tricky issues of syntactic ambiguity reduction and word sense disambiguation as well as providing innovative and original solutions for constituent generation and constraints management. To provide an insight into how it operates, the system's general architecture and components, as well as its lexical, syntactic and semantic resources are described. After that, the paper addresses the mechanism that performs selective parsing through an interaction between syntactic and semantic information, leading the parser to a coherent and accurate interpretation of the input text.

Keywords: syntax, semantics, NLP, parser, parsing strategies, word sense disambiguation, lexical database, semantic net.

## 1. SYNTAGMA's architecture

This paper discusses SYNTAGMA, a rule based NLP system addressing the tricky issues of syntactic ambiguity reduction and word sense disambiguation as well as providing innovative and original solutions for constituent generation and constraints management. The first section addresses its general architecture and components to provide an insight into how it works. The second section describes the system's lexical, syntactic and semantic resources. The last section shows how the system performs selective parsing through an interaction between syntactic and semantic information, leading the parser to reach the correct interpretation of the input text. A schema of the whole architecture is given in the appendix.

SYNTAGMA is the result of research into parsing strategies started by the author in 1989. The development of the current architecture began in 2011 after some experiments in data driven dependency parsing, which led the author to conceive a radically new parsing mechanism. Since its theoretical background has been discussed in a previous article[1], the focus here is on its architecture, components and resources.

The parser's core engine is language independent. All language specific rules and data are defined at the level of its lexical and syntactic resource files. Current implementation uses lexical, syntactic and semantic information which has been extracted automatically from an Italian dictionary, but these data are linked to the lexicon and the semantic net of other languages[2].

The system follows a bottom-up deterministic constituency parsing method, starting from the output of the Part of Speech Tagger (PoS-Tagger), which is a list of terminal categories, and progressively builds more and more complex structures until it reaches the highest constituency level the given input text allows. At the end of the process, constituency parse trees are transformed into dependency trees. Both output formats are available.

The system's behavior can be modulated by adjusting parameters, including the target language, the linguistic register (which can allow or prohibit some types of linguistic structures) and the strength of the selection mechanism during the parsing process (CSBS *Constituent-Selection-By-Score*).

---

[1]  D. Christen, *Syntagma. A Linguistic Approach to Parsing* (2013). See this article for references and related bibliography as well.

[2]  SINTAGMA's lexical database entries have been linked to WordNet *synsets* using WSD techniques.

The main components of the system are the following:

a) The *PoS-Tagger* (Part of Speech Tagger), which includes a tokenizer and performs Named Entities Recognition as well.

b) The *Constituent Generator*, which works recursively bottom-up using its own output (registered in the *Constituent Stack* during a previous cycle) as input for each new generation cycle. Starting with terminal categories, cycle by cycle, more complex structures are built by putting together the constituents generated from segments of the input string. A constituent record contains all the relevant lexical and structural information of a constituent and its sub-constituents. Constituent records stored in the stack have to pass the filters before they can be used in the successive cycle. Filters can inhibit a constituent's later use.
A component called *Ambiguity Solver* assigns a score to lexically or structurally ambiguous constituents. If the parser works with the parameter CSBS set to ON, only those concurrent constituents (competitors) reaching the highest score will be considered by the *Constituent Generator* for building new structures. The process ends if no new constituent record has been generated at the end of a parsing cycle.

c) The *Constituent Stack* is an array where all generated constituents' records are stored when they have successfully passed the filters. At the beginning of the parsing process, the stack contains only the output of the PoS-Tagger, the terminal categories. During each cycle its content grows as the *Constituent Generator* progressively delivers the constituents it builds. Constituents stored in the stack can be inhibited by filters and become inaccessible for the *Constituent Generator*.

d) After each cycle the *Filter Module* performs a selection of constituent records stored in the *Constituent Stack* by means of specific filter components dedicated to different aspects of grammar. The most important filter components are the *Argument Structure Filter*, the *Constraints Filter* and the *Co-Reference Filter*:

   d1) the *Argument Structure Filter* verifies the congruency between a given constituent pattern (the sequence of its sub-constituents) and the valency of the constituent head, according to the *projection principle*[3].

   d2) the *Constraints Filter* checks whether all constraints are satisfied. Constraints can arise at different levels, depending on where they were declared. They can come from the lexical item, from a given word sense or from a constituent pattern. Some constraints are built-in, like NP internal agreement and subject-verb agreement.

   d3) the *Co-Reference Filter* ensures the saturation of co-reference indices if a constituent contains traces or gaps. Empty elements without antecedent are allowed in intermediate constituents, but have to be defined within the following parsing process cycle, otherwise the constituent will be inhibited. This aspect is checked through different tools which respectively address interrogative and relative clauses, non-overt subjects and traces in subordinate clauses and various types of ellipsis in coordination.

e) The *Ambiguity Solver*. In our normal linguistic activity we are not aware of how frequent lexical and structural homonyms are because we automatically apply a wide range of criteria to select the most probable interpretation of an input. These criteria consist principally in semantic information about the input text and in our general knowledge of the world or about a specific domain. Other criteria concern pragmatics, communication context, text type, prosody and, finally, linguistic habits involving a probability calculus. Some of them are easier to formalize than others: the current version of SYNTAGMA

---

3    See Section 3 of this paper.

includes rules that manage the most relevant of these parameters.

Each constituent record stored in the stack is compared to the others. If they share some relevant properties they become competitors. Thus they need to be subjected to the tests contained in a module called *Ambiguity Solver*, which deals with ambiguities resulting from parsing process. Tests involve several aspects mentioned above and points are assigned to the competitor matching a given feature. Once the tests have been carried out, the score of each constituent is calculated. As said, the score can act selectively for the *Constituent Generator* if the CSBS parameter is on. This enhances the economy of sentence processing, reducing the number of parse trees available in the stack at the beginning of a new cycle.

f) The *Anaphora Solver* performs co-reference indexation for PRO subjects, pronouns and possessives. It operates across sentences as well, searching for anaphora antecedents in previous sentences too.

## 2. Resources

The main resources of the system are language specific. They are the *Lexicon*, the *Constituent Pattern List*, the *Argument Structure* database and the *Semantic* database. In the current version of SYNTAGMA they are available for Italian only. However further versions of the system aim to be multi-lingual. All data has been extracted from the same source dictionary, which ensures the internal coherence of the whole database, providing a strong correlation between the morphological, syntactic and semantic data of all lexical entries.

Beside the lexical database there is a *Named Entities* database that comes from *DBpedia*. It is employed first of all by the PoS-Tagger. It contains semantic tags (like *person*, *place*, *org*), which are used by the *Filter* module and by the *Ambiguity Solver* in order to evaluate the semantic consistency (i.e. the internal congruency) of a given interpretation of the input string.

### 2.1 Syntagma Lexical Database

The data that constitute the *Syntagma Lexical Database* (SLD) were all extracted through automatic procedures from the same source dictionary, originally in HTLM format[4].

The SLD contains a main table, *Lemma* (headword), in which the lexical entries with their grammatical features are listed. It contains links to aliases or alternative forms as well. Word specific constraints can be registered at this level. Each lexical entry is identified by a specific value in the index field LEX_ID. The *Forms* table lists all inflectional forms and their morphological features. Table *Forms* is related to table *Lemma* through the index field LEX_ID and *Forms* is used by the PoS-Tagger in order to identify the words in the input string. Word senses (i.e. their definition and semantic properties, links to their synonyms and to the domains they belong to) are collected in a table called *Meanings*, with the index field MNG. A fourth table, the *Argument Structure* table, is used during syntactic analysis by the *Argument Structure Filter*. This table contains the word's argument structure (*valency* in Tesnière's terminology), which is specific to a given sense of lexical entries. The relationship of *Lemma* records to *Argument Structure* records is one-to-many, because a lexical item can have more than one sense, each with its own valency. The index field of the *Argument Structure* table is MNG (meaning). For each MNG (i.e. word sense) the members of its argument structure are listed.

---

## 2.2  Constituent Patterns List

### 2.2.1 The structure of constituent patterns

SYNTAGMA's syntactic resource is the *Constituent Patterns List*. Syntactic constituents are listed in a table (a csv file, easy to edit and manage). They are described according to structural and generative grammar tradition in terms of their derivative and recursive properties. But in contrast to this tradition, they address the "surface" level of linguistic utterances. The relationship between the surface-level and its corresponding "deep structure"/"stemma"/"dependency tree"/"semantic representation" (whatever name we choose for the result of a parsing process) is fundamentally asymmetrical. The surface-level is linear (one dimensional), the deep structure implies vertical (two-dimensional) relations as well; in the former some information can be left unexpressed, in the latter these ellipses need to be filled in order to return a complete representation of what a people normally understand once they have processed the input.

Since the starting point of the parsing process is at the linear-level of the input string, a more appropriate term for these structures is *constituent patterns* or simply *patterns* rather than the traditional description of syntactic constituents.

The range of syntactic structures the parser is able to manage is as wide as the list of constituent patterns. The current Italian version lists about 600 patterns (including split sentences, left or right dislocation, i.e. topicalization, phrase and clause coordination with ellipsis), which allow the system to give the correct interpretation of a very wide spectrum of Italian sentences and text types. Constituent patterns are language specific and therefore need to be edited manually for each language the system will analyse.

The file containing these patterns is hierarchical. First is a list of the terminal categories (noun, verb, article, adjective, adverb etc.). After that come the constituents resulting from a combination of other constituents.

Special status is given to intermediate constituents containing empty categories like traces and gaps.

Empty categories are necessary for a constituent to satisfy the argument structure of the constituent head and to pass through the *Argument Structure Filter* successfully. They occur typically as constituents of interrogative, relative and subordinate clauses. Intermediate patterns containing empty categories can become part of coordinative structures as well[5]. These intermediate patterns should preferably precede the constituents they can become part of in the *Constituent Patterns List*. This enhances the economy of the parsing process, because they can be deleted from the *Constituent Stack* at the end of the cycle, since by that point they should have been absorbed by a higher level constituent.

Each constituent pattern consists of its constituent label (i.e. phrase marker or tag) followed by a series of one to a maximum of four datasets in brackets which contain:

    I)    The sequence of sub-constituent tags (*sub-constituent* dataset)
    II)   The dependency relations between the sub-constituents (*dependency* dataset)
    III)  The syntactic function of the sub-constituents (*syntactic function* dataset)
    IV)  The constraints over sub-constituents (*constraints* dataset)

The following examples show some low-level *constituent patterns*:

    Det, (611)
    Det, (612)
    Pron, (550)
    Pron, (560), {C(TAG=C) T(CAT=560) R(FNCT=obj)}
    N, (noun)
    Adj,(400)
    AdvQ, (310)

---

[5]   More about this in: *Coordination: patterns, rules and procedures* (2015).

The pattern is recognized by its tag, i.e. phrase-marker, and is followed by the *sub-constituent* dataset with, in these examples, only one element; so no *dependency set* is required. In this example, the subcategories are all terminal elements whose labels refer to some category in the lexical database. Category 600 is articles: 611 refers to definite and 612 to indefinite articles; pronouns belong to category 500 and include many subclasses in the Closed Classes section of the lexical database. No function set is needed because function will be assigned to these constituents at a higher level. But they can have a constraint expression in their constituent pattern, as seen with the category 560 Pron pattern. Constituent AdvQ has quantitative adverbs of class 310 as sub-constituents. Thus, this constituent can become a sub-constituent of a complex adjectival phrase (AdjP):

AdjP, (AdvQ, AdjP), (2,0)

The *dependency set* (referring to the position of sub-constituents in the pattern) tells us that the AdvQ depends on the second element in the sequence, that is the AdjP. It corresponds to the constituent's dependency tree. The next example shows a NP pattern corresponding to an input like "this beautiful place":

NP,  (Det, AdjP, N), (3, 3, 0),  (det, mod, 0)

The pattern tag is NP. This is followed by three data sets: the *sub-constituent dataset*, which tells the *Constituent Generator* that the input has to contain a determiner (Det), an adjectival phrase (which can also consist in more than one coordinated adjective), followed by a noun (N). The numbers 3,3,0 in the *dependency set* show that both the determiner and the adjective are dependent on the third element in the sequence, so this becomes the constituent head and its own dependency value is set to zero. The third dataset defines the syntactic function of each element of the sequence: the first is the determiner (*det*), the second a modifier (*mod*) and the head has no function assignment because it will receive it the moment this NP takes its place in a higher ordered constituent pattern: it can for example become the subject or the object of a clause. No constraints are mentioned for this pattern. The empty categories, *pro* (null-subject), *trace* and *gap*, are terminal constituents.

### 2.2.2 Constraint expressions

Constraint expressions are a sequence of units, each unit consisting of the *unit-functor* and three arguments: *tag*, *operator* and *value*:

Functor(Tag Operator Value)

Depending on the symbol, the *functor* indicates the context (*C*) within which a constraint becomes active or its target (*T*) or the content of a restriction (*R*). Other types of functor can be defined as well, such as level (*L*), direction (*DIR*) and distance (*DIST*). The arguments of a functor are: the *tag*, which can be a position in the constituent pattern, a constituent label, a syntactic function or some grammatical category; the *operator*, which may be = (equal) or # (not-equal); and finally the *value* that the target has to match.
Constraint units and constraint expressions can be combined through the logical operators AND/OR. So one can write complex restrictions in the form of logical expressions, which are solved by a dedicated module of the system and then sent to the *Constraints Filter*, which then verifies whether the grammatical features of the relevant target element match the constraint.
In the following example the constraint expression is related to a simple, non-finite clause pattern:

CP,(NP,V),(2,0),(subj,v), {AND(*T*(CT=1) *R*(CAT=pro), *T*(CT=2) *R*(VMD=non-finite))}

It tells the parser that the subject NP has to be a *pro* and that the V must have a non-finite verbal mood. The target functor (T) uses the parameter CT, which indicates the position of the affected constituent in the pattern.

The next example shows a constraint expression addressing the determiner of a NP, forcing it to be singular if the semantic feature of the head-noun is "mass" (the noun's semantic feature must be different from "mass" or the determiner has to be singular):

OR(*T*(TAG=N) *R*(SEM#mass), *T*(FNCT=det) *R*(NUM=sing))

Constraints can be asserted at three different levels: a) at the lexical level they will impact all occurrences of a given word; b) at the word sense (MNG) level they affect only some of the word's meanings; c) at the pattern level they can address all kinds of sub-constituent properties. Therefore there are constraints which cannot be checked at the same level at which they are declared. In these cases the *context functor* (*C*) will be used. When a constituent record is copied into a higher level constituent, it keeps its constraints and they will be checked only at the moment the current constituent tag matches the value requested by the *context functor*. There are, for example, some pronouns (elements belonging to category 560 in the closed classes database), whose syntactic function is restricted to the object role when they occur in a clause. So the constraint that comes with this lexical item can be verified only when it arises at a clause level *C*(TAG=CP), not earlier, otherwise a NP constituent would never be generated with category 560 elements:

*C*(TAG=CP) *T*(CAT=560) *R*(FNCT=obj)

During the parsing process, this constraint remains on "stand-by" until the context value matches that of the constituent containing the restricted term, that is: when the constituent tag corresponds to CP (clause phrase). The moment this occurs, the expression is passed to the *Constraints Filter*, which checks whether the target element, identified by category CAT=560, satisfies the restriction that requires it to have the object function in the given clause.

### 2.2.3 Pro, Traces and Gaps

*Pro*, *trace* and *gap* are terminal categories, which can form NP, PP and CP constituents[6]. *Pro* is a non-overt subject that typically appears in non-finite clauses depending on a subject-control verb, in interrogative clauses, in relative clauses and in adjunct clauses. It can also correspond to PRO subject (which are normal in Italian sentences). *Traces* are used in interrogative, relative and subordinate clauses. Patterns containing *gaps* are used in coordinated ellipsis structures when some argument is lacking in the first clause but expressed in the coordinated clause. Ellipsis can affect the coordinated clause as well, for example in VP ellipsis. Empty constituents are necessary in order to allow word senses, whose argument structure needs the missing argument (according to the *projection principle*) to pass the *Argument Structure Filter*.

The analysis of a sentence like: "Whom do you think John wants to invite?" uses the following clause patterns, some of which contain traces:

(1)   CP, (NP, V, NP), (2,0,2), (subj,v,obj)

This corresponds to the deepest subordinate clause, that is a subject-verb-object clause in which NPs are both left unexpressed ("*pro* invite *trace*"). Pattern (2), below, is an English-specific, non-finite clause pattern, which needs the particle *to* (a terminal category) as its first constituent, and a clause as second element, a constraint which affects the verbal mood: R(VMD=non-finite):

(2)   CP, (To, CP), (2,0), (to,0), {T(CT=2) R(MDV=non-finite)}

---

[6]   Related paper: *Parsing solutions for pro, traces and ellipsis* (2015), where the relevant principles of SYNTAGMA's grammatical framework are discussed as well. Some aspects of these principles may diverge from grammars on which current NLP research is based. The theoretical references and the reasons for these divergences are examined in the quoted paper, too.

The next pattern (3) has a NP as subject and a clause (CP) as object, both dependent on the verbal head. It also contains an intermediate *trace*, dependent on the subordinate clause (that is the fourth element in the sequence)[7]. The function *robj* is assigned to the trace, which is an antecedent-function of the object trace contained in the subordinate clause. The constraints address only the subordinate clause, a target (T) identified by its position in the sequence (CT=4). There are three restrictions this target has to satisfy: its verbal mood has to be non-finite, its subject has to be the *pro* and its object needs to be a *trace*.

(3)   CP, (NP, V, Trace, CP), (2,0,4,2), (subj, v, robj, obj), {T(CT=4) R(VDM=non-finite), R(SUBJ=pro), R(OBJ=trace)}

In our example, the segment "to [*pro* invite *trace*]", corresponding to pattern (2), can become the object clause of the intermediate constituent pattern (3):

[$_3$ John wants *trace* [$_2$ to [$_1$ *pro* invite *trace*]]]

In order to complete the analysis, a fourth pattern will be evoked, that is an interrogative clause:

(4)   CP, (PronInt, Do, NP, V, CP), (5,4,4,0,4), (robj, do, subj, v, obj)

Its first constituent is an interrogative pronoun with an object-antecedent function (*robj*), dependent on the fifth element of the sequence, the subordinate clause, which is the object of the head-verb (in our example "think"). The second constituent is the *do* particle, used in English interrogative clauses, which agrees with the subject (a built-in grammar constraint). Pattern (3) is a typical intermediate pattern: it will not be kept in the *Constituent Stack* at the end of the cycle that generated it. Instead it has to be absorbed by a higher level constituent (like pattern 4) within the same cycle.

In our example, the analysis of the input sentence is successfully completed and the *Co-Reference* module can populate the reference indices of the interrogative pronoun and of the traces as follows:

[$_4$ Whom$_i$ do you think [$_3$ John$_j$ wants *trace*$_i$ [$_2$ to [$_1$ *pro*$_j$ invite *trace*$_i$]]]]

This brief excursion into SYNTAGMA's syntax cannot be exhaustive and is intended as an introduction to the more detailed papers mentioned in the footnotes.

## 2.3  The Argument Structure Table

As we have seen (2.1), argument structures are registered in a specific table which is part of the lexical db[8]. They are word sense ordered and linked to semantic information through the index MNG (meaning). If a word sense carries an argument structure, its members appear as records of this table. Each record consists of the syntactic function label (*subj, obj, iobj* etc.) and the constituent tag (NP, PP, CP, etc.), which are essentially the main features of an argument's projection. If a connective is needed, its lexical form can be defined in a dedicated field, which will be checked by the *Argument Structure Filter*. Beside this, these records contain all the syntactic information affecting the argument structure of a specific word sense that could be extracted from the source dictionary. It can include positional, morphological, lexical and semantic constraints over argument projections. These constraints are checked by the *Constraints Filter*. Every element of the argument structure must have a corresponding sub-constituent in the pattern in order to satisfy the requirements for the correct projection of the constituent head according to Generative Grammar's *projection principle*.

---

[7] Trace-antecedent governors need to be assigned to the head of a subordinate constituent for intermediate and argument traces. This rule is defined when writing the pattern's dependency set and affects overt antecedents like relative, interrogative and clitic pronouns as well intermediate traces. It therefore leads the *Co-Reference Filter* to search for the antecedent of some trace in the immediately superordinate syntactic context. Co-reference chains can therefore easily be established. This general principle unifies in one process the handling of apparently different kinds of phenomena.

[8]   Examples of argument structures are given in Section 3.

The *Argument Structure* can even assign semantic values to an argument enabling it to predict that its projection will be a constituent with a given semantic feature (for example a NP labeled as "PERSON"). This plays a fundamental role during the filtering process done by the *Constraints Filter* and by the *Ambiguity Solver*, because these modules use the interaction between syntactic data and semantic information in order to select congruent parse-trees and to reach, as far as possible, a non-ambiguous interpretation of the input.

## 2.4 The Semantic Network

The fourth resource employed by the parser is an automatically generated Semantic Network, starting from the same source dictionary as the lexical database (SLD). This ensures the internal coherence of the whole system's resource database since all entries are co-indexed.

In order to generate the semantic net, the dictionary glosses were first parsed; after that, the output of the parsing process was transformed into *semantic frames* by a dedicated module[9]. In the semantic net the semantic feature of word senses (identified by the MNG index) are formalized and reduced to "primitive" predicates such as *perception, cognition, speech-act, action, thing, person, animal, place*, and sets of hierarchically ordered "primitive" predicates, so that a verb like "advance" is defined by the set: *(action(change(place(forward))))*, resulting from the processed lexical gloss of this verb (without human/manual editing intervention). Semantic relations expressed in different forms in dictionary glosses have been subjected to the same process and were formalized as: *token_of, part_of, has_part, has_quality, has_agent, has_object, has_cause, instrument_of*. From these the respective inverse relation may be deduced (*has_agent>agent_of; has_cause>cause_of; ...*).

The semantic net is used by the parser in disambiguation tasks and in some tests that contribute to assigning a score to a given interpretation (i.e. the parsing process result) of the input.

## 3. Selective Parsing through Semantic and Syntax Interaction

In this last section I want to show how SYNTAGMA performs selective parsing, which affects both parse-trees and word senses through a continuous interaction between the syntactic and the semantic information accessible at a certain stage of the parsing process.

The constituent generation process follows the *Constituent Patterns List* order and builds structures by putting together the elements that are currently stored in the *Constituent Stack*, regardless of their grammatical features. They only have to match the requested sub-constituent label and satisfy the *contiguity constraint*: that is they have to respect the word order of the input string[10]. While the *Constituent Generator* ensures only the congruency between some sub-list of the input string and a constituent pattern, the *Argument Structure Filter* checks whether the selected patterns match the argument structure of the constituent head, while the *Constraints Filter* verifies whether terminal constituents satisfy morphological and semantic constraints, which are asserted at the lexical or the word sense level. This is the kind of selection which affects the syntactic level: patterns are only kept in the *Constituent Stack* if they can be considered a projection of the valency of a given word's sense (MNG); otherwise they are rejected. Since argument structures derive from the semantic structure of words, we have here a selection mechanism that starts from the semantic level and ends at the syntactic level.

---

[9]   For the discussion of this operation the reader is referred to: *Rule-Based Semantic Tagging. An Application Undergoing Dictionary Glosses* (2013).

[10]  How empty (non-overt) categories can be integrated without violating the contiguity constraint is explained in my paper: *Parsing solutions for pro, traces and ellipsis* (2015).

But at the same time the system performs a selection that impacts the semantic level as well, using information from the syntactic level (i.e. from the constituent patterns).

At the beginning of the parsing process, every lexical entry receives its complete MNG index (i.e. word sense references). So at this stage a word can potentially have all the meanings the source dictionary has assigned to it. When a lexical entity is copied into a constituent record, a copy of its MNG index is made as well. So the MNG index-set becomes part of a *Constituent Stack* record's lexical information. Since argument structure is word-sense related, only those MNGs will be accepted whose features match the constituent pattern that could have been generated from a given segment of the input string. During the parsing process, the *argument structure filter* deletes the MNG references (i.e. word senses) from the MNG set whose argument structure doesn't match the given pattern structure. The *Constraint Filter* performs the same selection if pattern related constraints cannot be satisfied by some word-sense related feature. Consequently the MNG set of the constituent lexical entries becomes progressively smaller during the parsing cycles, which means that the word's semantic spectrum becomes more defined and precise. In this way, structural selection leads automatically to word-sense selection.

The following example shows three different (beside many others) meanings of the verb "amare" ("love") with their valency. Constraints and semantic features are registered with the related valency in the *Argument Structure* table:

MNG 3085.01,  LEX "amare",   PRF "Provare amore e affetto verso qlcu."

| ARG | FNCT | CAT | CONN | CTRL | SEM |
|-----|------|-----|------|------|-----|
| 1;1 | subj | NP | NIL | NIL | NIL |
| 2;1 | v | VP | NIL | NIL | NIL |
| 3;1 | obj | NP | NIL | NIL | NIL |

Word sense 3085.01 shows a transitive argument structure. It means: "to feel love and affection for somebody". The third argument, the object, has to be a NP. The *Control* parameter has the default value "NIL".

Conversely, beside showing an NP projection, word-senses 3085.08 and 3085.09 also show CP projections. But their meanings are different and they assign different constraints to their object clause:

MNG 3085.08; LEX "amare";    PRF "provare diletto nel fare qlco., ricevere diletto da qlco.,";

| ARG | FNCT | CAT | CONN | VMD | CTRL | SEM |
|-----|------|-----|------|-----|------|-----|
| 1;1 | subj | NP | NIL | NIL | NIL | NIL |
| 2;1 | v | VP | NIL | NIL | True | NIL |
| 3;1 | obj | CP | NIL | non-finite | NIL | "GOAL" (EVENT) |
| 3;2 | obj | CP | "di" | non-finite | NIL | "GOAL" (EVENT) |
| 3;3 | obj | CP | "che" | subjunctive | NIL | "GOAL" (EVENT) |

MNG 3085.09; LEX "amare";        PRF "desiderare, volere qlc."";

| ARG | FNCT | CAT | CONN | VMD | CTRL | SEM |
|-----|------|-----|------|-----|------|-----|
| 1;1 | subj | NP | NIL | NIL | NIL | NIL |
| 2;1 | v | VP | NIL | conditional | False | NIL |
| 3;1 | obj | CP | "che" | subjunctive | NIL | "GOAL" (EVENT) |

The word sense 3085.08 can be translated as "to like to do something" and has three possible configurations for its object argument. The first one is a non-finite clause (without conjunction). Since the control parameter (CTRL) is set to true, the *pro* subject of the non-finite clause will be co-referent with the head-verb subject (1;1). The second is a finite clause introduced by the conjunction "che" ("that"), which permits an overt subject and does not have to be co-referent with the main subject (1;1) since the control feature is set to false. Even if the subject of the subordinate clause were left unexpressed, that is a PRO, there would be no co-reference with the subject (1;1). The verbal mood of this subordinate clause has to be subjunctive. The third projection is a non-finite clause introduced by the conjunction "di" ("of"), whose *pro* subject is controlled by the verb "amare" (2;1) and is therefore co-referent with the subject (1;1). The subordinate clause can also be affected by negation, as in: "Paolo non ama ballare" or "Non amava più di muoversi, di vedere la luce" (G. Deledda)[11].

Word sense 3085.09 is optative and can be translated as "to want something (to happen), to desire something". The object has only one projection: a finite clause introduced by the conjunction "che", whose subject is not co-referent with the subject (1;1). The conditional mood is imposed on the main clause verbal head, while the subordinate clause's head should be in the subjunctive verbal mood.

Both word senses assign the semantic feature "ACTION" to their object.

Depending on the input sentence and the constituent pattern that can be generated from it, one of these three word senses will be kept and the others will be deleted from the MNG index set of the verb "amare" in the given clause constituent, as the following examples show:

a)       "Paolo ama l'insalata"    ("Paul likes salad")
b)       "Paolo ama cantare"       ("Paul likes to sing")
c)       "Paolo ama che gli amici lo lodino" ("Paul likes to be praised by his friends")
d)       "Paolo amava di cantare ogni giorno per qualche ora" ("Paul liked to sing for several hours every day")
e)       "Paolo amerebbe che gli dicessero la verità" ("Paul would like them to tell him the truth")

The *Argument Structure Filter* can establish the following correlations between patterns and word senses:

a) matches the pattern CP(NP,V,NP), that is a projection of word sense MNG 3085.01;
b) matches the pattern CP(NP,V,CP),{T(CT=3) R(VMD=non-finite)} a projection of  MNG 3085.08;
d) matches CP(NP,V,CP),{T(CT=3) R(VMD=non-finite), R(conn="di")}, also a projection of MNG 3085.08;
c) and e) match the pattern CP(NP,V,CP),{T(CT=3) R(VMD=subjunctive), R(conn="che")}, which is a projection of both MNG 3085.08 and MNG 3085.09. Disambiguation can be performed by the *Constraint Filter* of the main verb's mood, which is indicative for word sense 3085.08 and subjunctive for word sense 3085.09.

At the beginning of the parsing process, at the V (verb) constituent level, the MNG set of the verb "amare" is [3085.01, 3085.02,..., 3085.08,3085.09], which comes with the terminal constituent identified by the PoS-Tagger. Once the verb "amare" reaches the CP level, the *Argument Structure Filter* and the *Constraints Filter*

---

[11]    This last non-finite form, introduced by a preposition, is rare and obsolete but documented. Translation: "Paolo doesn't like dancing", "He didn't like to move any more, nor to see daylight"

(see section 1) perform a selection, after which only its congruent MNG values will be kept in the set. Beside the two meaning selection mechanisms discussed above, a third type of WSD is performed by the Ambiguity Solver component (Section 1 e). The Ambiguity solver compares the semantic features assigned to a given element of the argument structure and those of the lexical item, which becomes the projection of this argument. This is easily demonstrated through the following examples related to the same verb "amare".

While MNG 3085.01 (above) requires a "PERSON" as its object ("per qualcuno": "for somebody"), the source-dictionary definition of MNG 3085.05 and 3085.06 (below) says that their objects are THINGS ("per qualcosa": "for something"). This information has been automatically formalized (during the lexical database generation) and put in the argument structure as follows:

MNG 3085.05; LEX amare; AUX "avere"; TRN 2; RFL 0; CTRL T; VL "[subj-v-arg]";
    DEF "Sentire affetto, attaccamento per qlco.";
    EX "amare il proprio paese, il proprio lavoro";

    [1;1; FNCT subj;   VCAT NP;    CONN ; OPT F;  VSEM NIL;            VREF; VRESTR ;]
    [2;1; FNCT v;      VCAT VP;    CONN ; OPT F;  VSEMNIL;             VREF; VRESTR ;]
    [3;1; FNCT obj;    VCAT NP;    CONN ; OPT F;  VSEM "THING";       VREF; VRESTR ;]

MNG 3085.06; LEX amare; AUX "avere"; TRN 2; RFL 0; CTRL T; VL "[subj-v-arg]";
    DEF "sentire inclinazione, interesse, attrazione per qlco.";
    EX "amare lo studio, la musica, i viaggi, il denaro";

    [1;1; FNCT subj; VCAT NP;      CONN ; OPT F;  VSEM NI;             VREF; VRESTR ;]
    [2;1; FNCT v;   VCAT VP;      CONN ; OPT F;  VSEM NIL;            VREF; VRESTR ;]
    [3;1; FNCT obj; VCAT NP;      CONN ; OPT F;  VSEM "THING";       VREF; VRESTR ;]

In a sentence like "Paul loves his mother", since "mother" is semantically tagged as "PERSON" in SYNTAGMA's semantic net, the *Ambiguity solver* selects MNG 3085.01 and deletes MNG 3085.05 and 3085.06 from the MNG index of the constituent head verb. Conversely it keeps both MNG 3085.05 and 3085.06, and deletes 3085.01 if the lexical entry corresponding to the object has a tag that doesn't match "PERSON", like "job", "country", "music" or "money". Since THING is non-specific (i.e. more general) compared with PERSON or ANIMAL, if the object is the word "horse", tagged as ANIMAL, the *Ambiguity Solver* will exclude MNG 3085.01 and choose both 3085.05 and 3085.06.

Some dictionary definitions of the same verb assign a specific semantic tag to the subject, like the following:

MNG 3085.09; DACZ 3; LEX amare; AUX "avere"; TRN 2; RFL 0; CTRL T; VL "[subj-v-arg]";
    DEF "Detto di animali, prediligere qlco.";
    EX ("i gatti amano la solitudine");

    [1;1; FNCT subj; CAT NP; CONN ; OPT F;  VSEM "ANIMAL";       VREF; VRESTR ;]
    [2;1; FNCT v;     CAT VP; CONN ; OPT F;  VSEM; NIL;           VREF; VRESTR ;]
    [3;1; FNCT obj;  CAT NP; CONN ; OPT F;  VSEM ("THING");      VREF; VRESTR ;]

MNG 3085.10; DACZ 3; LEX amare; AUX "avere"; TRN 2; RFL 0; CTRL T;  VL "[subj-v-arg]";
    DEF "detto di piante, abbisognare di qlco. per prosperare";
    EX ("gli agrumi amano il clima mediterraneo");

    [1;1; FNCT subj; CAT NP;      CONN ; OPT F;  VSEM "VEGETAL";      VREF; VRESTR ;]
    [2;1; FNCT v;     CAT VP;      CONN ; OPT F;  VSEM NIL;          VREF; VRESTR ;]
    [3;1; FNCT obj;  CAT NP;      CONN ; OPT F;  VSEM NIL;          VREF; VRESTR ;]

So the *Ambiguity Solver* will choose MNG 3085.09 if the input sentence is "Piante che amano un clima caldo umido prosperano vicino all'equatore" ("Plants that need a warm moist climate thrive close to the equator", where "amare" has a sense that is closer to "need" than to "like", since in the semantic net "pianta" is tagged, among many other word senses, as "VEGETAL", which matches the tag assigned to the subject in the argument structure.

And it will choose MNG 3085.10 if the sentence is "Anche se gli elefanti africani e asiatici sperimentano occasionalmente temperature relativamente basse per brevi periodi, amano i climi caldi" ("Even if African and Asian elephants may occasionally experience relatively low temperatures for short periods, they prefer warm climates"), where "amare" is interpreted as "preferire" ("prefer"), because the word "elephant" has the tag "ANIMAL" in the semantic net.

The strength of this WSD mechanism is a working parameter of the system that the user can set. Therefore, if these semantic parameters didn't match, the parsing process would still produce a result, which would be the term that reached the highest score by some other criteria as discussed elsewhere in this paper.

**4. Conclusions**

This paper discussed a currently fully operational NLP system, called SYNTAGMA, which features some innovative solutions involving constituent generation and constraints management as well as addressing problems associated with syntactic ambiguity reduction and word sense disambiguation. The description of the system's lexical, syntactic and semantic resources showed how these combine to create a coherent internal database, within which all these different types of data are related to each other. Finally it focused on how the parser operates and how syntactic structures and sense related features continuously interact, performing reciprocal selections at different levels, leading progressively towards a coherent and accurate interpretation of the input.